# A sequential transit network design algorithm with optimal learning under correlated beliefs

Gyugeun Yoon[1]*, Joseph Y.J. Chow[2]
[1]Department of Civil and Environmental Engineering, College of Engineering,
Seoul National University, Seoul, Republic of Korea
[2]C2SMARTER University Transportation Center, New York University Tandon School of Engineering, Brooklyn, NY, USA
*Corresponding author email: gyugeun.yoon@snu.ac.kr

## Abstract

Mobility service route design requires demand information to operate in a service region. Transit planners and operators can access various data sources including household travel survey data and mobile device location logs. However, when implementing a mobility system with emerging technologies, estimating demand becomes harder because of limited data resulting in uncertainty. This study proposes an artificial intelligence-driven algorithm that combines sequential transit network design with optimal learning to address the operation under limited data. An operator gradually expands its route system to avoid risks from inconsistency between designed routes and actual travel demand. At the same time, observed information is archived to update the knowledge that the operator currently uses. Three learning policies are compared within the algorithm: multi-armed bandit, knowledge gradient, and knowledge gradient with correlated beliefs. For validation, a new route system is designed on an artificial network based on public use microdata areas in New York City. Prior knowledge is reproduced from the regional household travel survey data. The results suggest that exploration considering correlations can achieve better performance compared to greedy choices in general. In future work, the problem may incorporate more complexities such as demand elasticity to travel time, no limitations to the number of transfers, and costs for expansion.

# 1. Introduction

Mobility service route design depends on identifying the demand level in the potential service region of a system. In conventional transportation planning, an operator would collect survey data from travelers in the region and use that to forecast the demand for the whole route design. For example, these include survey results from samples of the population (e.g. Regional Household Travel Survey (RHTS) conducted by New York Metropolitan Transportation Council (NYMTC)) (NYMTC, 2010). Forecast models are then possible if travelers are familiar with the costs and benefits of the proposed mode(s) and how they perform relative to existing modes in the region. But for emerging transit technologies, such as modular autonomous vehicle fleets (e.g. Guo et al., 2017; Caros and Chow, 2021), more elaborate feeder-trunk routes that make use of various shared mobility options (e.g. Ma et al., 2019), or shared autonomous vehicle fleets that have little to no data to begin with, each new city deployment requires starting from scratch (Allahviranloo and Chow, 2019). Whereas regional travel surveys may be conducted every five or ten years, emergent technologies are unfolding on a yearly time horizon. Not having experienced an emerging technology mode, travelers' responses to hypothetical questions in stated preference surveys may not be as reliable when asked about them (see Bunch et al., 1993; Watson et al., 2020).

Instead, a mobility service system can accumulate its own data generated during the operation. These data are not limited to basic operation logs like vehicle trajectory and load profile data, but also cover interactions with passengers such as revealed preferences, ridership, average wait time, and ratings. In this way, the routes served by an operator become important sensors for collecting data. An operator can use these data to design their route system to fit the prevailing demand in a sequential or phased implementation (Yoon and Chow, 2020) instead of implementing a single design at once. In this context, service design should be implemented with explicit consideration of the use of the routes as sensors, i.e. placement of one route design versus another impacts not only service provided to travelers, but also in the knowledge gained from the more reliable data on those operated routes. This explicit consideration of route design as sensors is called "transit service route design with optimal learning."

In optimal learning, sequential decisions are made such that the information gained from each decision epoch is optimized to reduce the uncertainty for subsequent decisions. It is a type of reinforcement learning (RL) (Powell, 2021). Motivated by the similarity between sequential mobility route design and optimal learning, this study seeks a new approach to design mobility service routes tailored to local demand patterns by adapting a learning-based scheme. While earlier work on this problem from Yoon and Chow (2020) considered the use of contextual bandit algorithms sequentially building out one complete route at a time, that method assumes that knowledge gained from each route expansion is independent from other options. In a network setting, this is clearly not the case. An alternative approach is the knowledge gradient (Frazier et al., 2008), which does not require this assumption of independence between options. However, implementation of knowledge gradients in network settings can be problematic because of the way the covariance matrix between different route elements scales up (Ryzhov et al., 2012).

In the context of transit networks, we consider the problem of sequentially selecting a segment or route to expand under uncertainty in which the demand distributions are not known in advance. The decisions made each epoch seek to maximize the cumulative reward by considering the observed data sampled from the new service so that the subsequent decisions are better informed. A more formal definition of this problem is provided in Section 3. We propose several contributions to address it.

First, a system design is proposed for integrating optimal learning with correlated stochastic variables for sequential transit network design *at the segment extension level*. The design includes the flow of data, required inputs and outputs, and integration of optimal learning methods into a route design algorithm. Second, three different learning policies are compared in two experiments: a 5-by-5 grid network with artificially generated data and a realistic scenario based on public use microdata from New York City (NYC). In the latter, two benchmarks are included in the comparison: a greedy algorithm and the Chow-Regan (CR) reference policy (Chow and Sayarshad, 2016). These computational experiments provide insights on the performance of sequential transit network design with optimal learning regarding the inclusion of exploration.

This paper is structured as follows. First, existing works about transit route design and learning policies are reviewed. Second, the problem is defined, and the proposed route design algorithmic framework is presented. Third, numerical experiments are designed for illustrating and testing the proposed algorithm under different learning methods, and results are discussed. Finally, conclusions and prospective future advances from this research are presented.

## 2. Literature review

The review in this section examines the conventional and dynamic transit service route design and explores some optimal learning policies potentially applicable to the proposed methodology.

### 2.1. Mobility service route design with fixed routes

#### 2.1.1. Conventional approaches with static demand

Transit route design is one of the components of a line planning problem (LPP), where the latter more broadly includes service frequency or timetable scheduling. Since this study aims to implement optimal learning to transit route design, our review and following research focuses on the route design part, excluding other system factor designs like frequency setting, similar to set covering transit design problems like in Current et al. (1985), Current et al. (1994), and Wu and Murray (2005). Conventional approaches establish a set of fixed routes that optimizes objective values based on the static demand information. Routes are not assumed to change once implemented, which makes the system design simpler and more efficient. This is highly applicable when the demand level is expected to be stable, or the spatiotemporal scope is sufficiently narrow to perceive demand as constant.

Methodologies for generating routes for static line planning either build them based on the network characteristics (e.g. Ngamchai and Lovell, 2003; Baaj and Mahmassani, 1995; Israeli and Ceder, 1995; Chakroborty and Wivedi, 2002; Zhao and Zeng, 2008; Iliopoulou and Kepaptsoglou, 2019) or choose from a predefined set of "feasible routes", i.e. routes that satisfy such criteria regarding geometrical or operational attributes as total length or mandatory visits to certain nodes (e.g. Ceder and Israeli, 1998; Pattnaik et al., 1998; Tom and Mohan, 2003; Fan and Machemehl, 2006, 2008; Cipriani et al., 2012a, 2012b; Schmid, 2014; Walteros et al., 2015; Owais et al., 2015). Numerous methodologies used for generating route sets include route construction heuristics (Ceder and Wilson, 1986), genetic algorithms (Chien et al., 2001), column-generation algorithms (Borndörfer et al., 2007), and adaptive neighborhood search metaheuristics (Canca et al., 2017).

Several studies investigated route design with variable or elastic demand. For instance, if the public transit demand is determined by a mode choice model with travel time attributes such as a logit model, the ridership will depend on the service level it provides. However, they assume the demand matrix and the model are known to the planner, making the ridership estimable (e.g. Fan and Machemehl, 2008; Lee and Vuchic, 2015; Yoo et al., 2010; Gallo et al., 2011; Zarrinmehr et al., 2016).

*2.1.2. Sequential design*

Evolutional models and incremental network design are two examples of adjusting an existing network. Evolutional models modify the network to respond to changes in the surrounding environment (e.g. Fang et al., 2018; Zhang et al., 2023). For instance, an evolutional model for supply chains makes them actively reflect external market demand and internal competition-cooperation. Operators add or remove nodes and/or links for each time step based on the calculated addition and deletion probabilities. On the other hand, incremental network design expands networks a step at a time over multiple time steps associated with sequentially available budgets (e.g. Kim et al., 2008; Baxter et al., 2014), forcing planners to choose one of the alternatives for each period. Although both approaches sequentially modify networks to achieve the best solutions, they are established on the assumption of deterministic demand at least within each time step.

In contrast, dynamic demand involves uncertainty that prohibits the precise prediction of demand. Due to its higher complexity, route design under uncertainty is not well studied. A few studies tackle LPPs incorporating route design with two-stage stochastic programs or robust optimization (An and Lo, 2015, 2016; Liang et al., 2019). One strategy for mitigating uncertainty over a time horizon is to consider the buildout over multiple stages and to adapt subsequent stages to prior outcomes (Chow and Regan, 2011). Staged development is a natural approach to transit networks (Mohammed et al., 2006; Li et al., 2015; Sun et al., 2017; Yu et al., 2019). This notion of flexibility leads to a sequential network design problem under uncertainty, a more complex category of Markov decision processes where interdependent decisions are made dynamically over multiple periods with information revealed over time (Chow and Sayarshad, 2016; Powell, 2007). Approximate dynamic programming (ADP) algorithms and RL are typically used to optimize such problems. Whereas ADP typically assumes the distribution of the uncertainty is known throughout the decision process and seeks actions to anticipate the future, RL includes uncertainty in the *belief* of the distribution and manage the decisions between exploitation (optimizing the system under the current belief of the distribution) and exploration (optimizing the updating of the belief of the distribution). There are methods that combine the anticipative actions of ADPs with the optimal learning aspects in RL (e.g. Powell and Ryzhov, 2012a; Ryzhov et al., 2019).

RL has been proposed for extending routes (Wei et al, 2020). Yoon and Chow (2020) proposed a similar approach at the route level that builds feasible routes in advance and include each route as options for sequential expansion. However, the route enumeration can be an obstacle to optimize the system. An alternative is to design a transit network sequentially one segment at a time assuming shorter epochs in which the system obtains feedback after each segment extension; this would avoid route set generation. *No literature has considered the sequential route design problem with segment-level decisions, let alone using RL to optimally learn the demand.*

**2.2. Reinforcement learning in sequential planning for transportation systems**

RL can help determine how to efficiently estimate the consequence of our action with the limited knowledge that we have. Operators can update this knowledge about the environment through a series of actions and observations. Each action corresponds to a reward that operators are interested in: radio channel availability (Liu and Zhao, 2010), click-through rate of news articles (Li et al., 2010), and reliability of a chosen path (Zhou et al., 2019). The basic concept of RL consists of exploring and exploiting. After the system accumulates knowledge from exploring various options, it determines the best option to choose based on the combination of the current reward and expected future value. Depending on the perspective of operators, they can balance the weight of both factors. While risk-takers may explore and aim for higher return, others may be satisfied with the already known benefit value. These matches between actions and known environment information form a learning policy. Different learning policies define their own evaluation measures.

Not surprisingly, some transportation domains have adopted the learning techniques to dynamically reflect real data collected from previous operations and predict the best action that systems may take in the next period. **TABLE 1** highlights some transportation-related examples that adopted RL. Zolfpour-Arokhlo et al. (2014) developed a route planning model using multi-agent RL in a Malaysian intercity road network to reduce travel time between cities. Zhou et al. (2019) applied a multi-armed bandit (MAB) algorithm to sequential departure time and path selection considering on-time arrival reliability. Huang et al. (2019) used the knowledge gradient for allocating delivery vehicles to sectors in a region while minimizing the expected operational cost. Römer et al. (2019) implemented a contextual bandit process to control charging demands of EVs by adjusting the price and recommending stations to users. Zhu and Modiano (2018) dealt with travel time delays on the network where delays were collected only in total along the path and those of individual links would be revealed if selected. Three alternative methods are discussed in more detail.

**TABLE 1 Examples of Learning Techniques Used in Transportation Domain**

| Subject | Approach | Information (I) and Action (A) |
|---|---|---|
| Intercity route planning (Zolfpour-Arokhlo et al., 2014) | Q-value based dynamic programming | I: travel time<br>A: route choice |
| Sequential reliable route selection (Zhou et al., 2019) | Multi-armed bandit | I: generalized travel time, reliability<br>A: route choice |
| Delivery vehicle allocation (Huang et al., 2019) | Knowledge gradient | I: operational cost curve<br>A: vehicle allocation to regions |
| Demand management of EV charging stations (Römer et al., 2019) | Contextual bandit | I: load on grid<br>A: charging price adjustment |
| Stochastic online shortest path routing (Zhu and Modiano, 2018) | Combinatorial bandit | I: end-to-end delay<br>A: path choice |

*2.2.1. Multi-armed bandit*

MAB involves recommending from a set of fixed, independent options (i.e. arms) over multiple periods to minimize regret $\rho_n$ resulting from the revealed rewards in each trial. Regret represents the difference between a maximum possible reward and the acquired reward as shown in Eq. (1). For a chosen arm $I_t$ at time step $t$, the observed reward is $Y_{I_t,t}$. Having $K \geq 2$ arms, $Y_{i,t}$ of arm $i \in [1, K]$ is assumed to follow an unknown distribution (Bubeck and Cesa-Bianchi, 2012).

In reality, detecting $\rho_n$ is nearly impossible due to the unobservability of simultaneous rewards across all arms.

$$\rho_n = \max_{i=1,\cdots,K} \sum_{t=1}^{n} Y_{i,t} - \sum_{t=1}^{n} Y_{I_t,t} \tag{1}$$

Considering a length of observations $T$, the initialization period $\tau \leq T$ is set to build initial knowledge about distributions of rewards. During $\tau$, choices on arms can be made randomly instead of following a certain policy. When $t > \tau$, the agent chooses the arm that balances between the highest expected reward in the current period and lowering the upper bound of possible regret across all periods. An algorithm called "Upper Confidence Bound (UCB)" limits the increase of total regret at a rate lower than $O(n)$, reducing the uncertainty of the total amount of regret (Auer et al., 2002). The drawback of this approach is that every arm is assumed to be independent of each other.

### 2.2.2. Knowledge gradient

Knowledge gradient (KG) is an optimal learning technique that sequentially decides on an action taking into account the potential knowledge that may be gained in shaping subsequent decisions (Powell and Ryzhov, 2012b). If $S^n$ is the state after the $n$-th measurement, $S^{n+1}(c)$ is the state after choosing $c$ from the choice set $\mathbb{C}$ after the $n$-th measurement, and $V^n(S^n)$ is the value of being in state $S^n$ after the $n$-th measurement. It seeks the option that can improve $V$ the most by estimating the expectation of the value after the $n$-th measurement in Eq. (2).

$$v_c^{KG,n} = \mathbb{E}[V^{n+1}(S^{n+1}(c)) - V^n(S^n)|S^n] \tag{2}$$

After estimating $v_c^{KG,n}$ for all $c$, the policy chooses the best option $C^{KG,n}$ that maximizes $v_c^{KG,n}$ in Eq. (3). The derivation of $v_c^{KG,n}$ is illustrated in Section 3.1.3.

$$C^{KG,n} = \arg\max_{c \in \mathbb{C}} v_c^{KG,n} \tag{3}$$

### 2.2.3. Knowledge gradient with correlated belief

KG with correlated beliefs (KGCB) is similar to KG but considers correlation among options (Powell and Ryzhov, 2012b). The expected value of choosing $c$ after the $n$-th measurement is estimated by Eq. (4).

$$C^{KG,n}(s) = \mathbb{E}\left[\max_i(\mu_i^n + \tilde{\sigma}_i(c^n)Z^{n+1})|S^n = s, c^n = c\right] \tag{4}$$

where $\mu_i^n$ is a belief about the mean of a measure at $n$, $\tilde{\sigma}_i(c)$ is the change in our belief about option $c$, $Z^{n+1}$ is a random variable regarding the difference between the observation and belief, $S^n$ is state of knowledge, and $c^n$ is the chosen option at $n$. Specifically, a correlation matrix $\Sigma^n$ affects $\tilde{\sigma}_i(c)$, providing information about potential influences of choosing $c^n$ on other options. Section 3.1.3 describes the detailed calculation.

**2.3. Research gap summary**

There has been little effort to embrace demand uncertainty to sequentially expand transit networks. Several studies developed sequential network design techniques and dynamic methods to be responsive to fluctuating demand or be aligned with partially available budget but assumed that the demand is deterministic. Other researchers tackled demand uncertainties within their network design. However, their research was not based on either sequential network expansion or dynamic demand, which this study considers.

Some transportation fields have actively adopted RL, especially in traffic flow, traffic signals, and transit assignment (Abdulhai et al., 2003; Walraven et al., 2016; Cats and West, 2020). There are some common points between sequential transit network design and optimal learning policies. First, actions (segment extensions) correspond to rewards (additional demand covered). Second, a series of observations is used to improve the accuracy of the expected reward estimation. Lastly, an action at the current time step affects future ones. The following sections propose how to model a sequential segment-level route design process with optimal learning.

## 3. Proposed Methodology

There is a need to clarify the concept of "segment extension" and "route expansion." Both terms imply the physical growth of a system, but we distinguish segment extension and route expansion as two different levels as shown in **Figure 1**. A segment extension appends a segment to a current route and elongates its length. On the contrary, a route expansion adds a new route to an existing route set and improves the coverage of the system. When considering transfers, multiple routes can also cover travel demand across more node pairs.

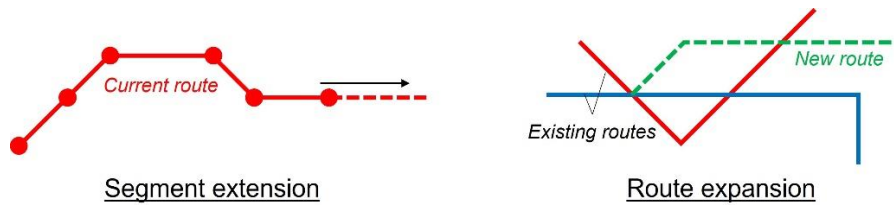

**Figure 1. Concept of segment extension and route expansion.**

Consider a system that is gradually expanded: we extend a route segment-by-segment and add it to the existing system. Each time a segment is added, some time passes in which new information is obtained from users served by that additional segment. The maximum length and number of routes are design criteria which can be derived from the budget availability.

This framework can be adapted to many types of transit systems with some modifications. For example, with mixed-traffic bus networks, the framework can be directly applied with a modification to allow prior route segments to be removed with a cost. For the case of fixed-guideway networks with high investment costs, the framework may not be reasonable to apply random pilots in the beginning due to the high cost of investment in rail. Instead, the prior would have to be obtained through stated preference surveys or alternative technology operations like buses initially to ascertain the demand patterns, after which the learning framework can be applied. In the case of on-demand microtransit or ridepooling, the decision variable would not be links and segments, but would be service region zones that would be served. This framework thus opens up new avenues of future research in sequential transit network design that can involve better design of priors using initial surveys or pilots, alternative network structures like zone-based service provision, integration of bus performance with traffic congestion, or networks that can grow and contract over time.

### 3.1. Problem Statement

**TABLE 2 Notations used in Section 3 and 4**

| Notation | Description | Notation | Description |
|---|---|---|---|
| $\pi$ | Learning policy | $G$ | Network |
| $N$ | Set of nodes | $n$ | Node |
| $A$ | Set of bidirectional segments | $a$ | Segment |
| $T$ | Time horizon | $t$ | Extension time step |
| $M$ | Number of segment extension trials | $\mathbb{N}(\cdot,\cdot)$ | Normal distribution with mean and covariance |
| $a^t$ | Segment that can be chosen at $t$ | $A^t$ | Set of $a^t$ |
| $a^{t,m}$ | $a^t$ chosen at segment extension trial $m$ | $N^t$ | Set of nodes corresponding to $A^t$ |
| $n^{t,m}$ | Node corresponding to $a^{t,m}$ | $v_{a^t}^{m,\pi}$ | Value of $a^t$ after $m$ trials under $\pi$ |
| $K$ | Number of candidate routes | $R_k$ | $k$-th route |
| $x_{i,j}$ | True OD demand between $i$ and $j$ | $X^t$ | Total OD demand coverage at $t$ |
| $X_{a^t}$ | Sum of all available $x_{i,j}$'s if appending $a^t$ | $\Delta X_{a^{t,M}}$ | Sum of additional $x_{i,j}$'s expected to be covered by appending $a^{t,M}$ |
| $a^{t(q)}$ | $q$-th segment of $A^t$ | $D^q$ | Set of possible $x_{i,j}$'s if integrating $a^{t(q)}$ |
| $\boldsymbol{\theta}^D$ | Vector of $x_{i,j}$'s | $\boldsymbol{\Sigma}^D$ | True covariance matrix of $x_{i,j}$'s |
| $\boldsymbol{\theta}^A$ | Vector of $X_{a^{t(q)}}$ | $\boldsymbol{\Sigma}^A$ | Covariance matrix of $X_{a^{t(q)}}$ |
| $\boldsymbol{\theta}_t^A$ | Vector of $X_{a^{t(q)}}$ at $t$ | $\boldsymbol{\Sigma}_t^A$ | Covariance matrix of $X_{a^{t(q)}}$ at $t$ |
| $\widehat{\boldsymbol{\theta}}_0^D$ | Initial prior of $\boldsymbol{\theta}^D$ | $\widehat{\boldsymbol{\Sigma}}_0^D$ | Initial prior of $\boldsymbol{\Sigma}^D$ |
| $\widehat{\boldsymbol{\theta}}_t^D$ | Updated prior of $\boldsymbol{\theta}^D$ at $t$ | $\widehat{\boldsymbol{\Sigma}}_t^D$ | Updated prior of $\boldsymbol{\Sigma}^D$ at $t$ |
| $\widehat{\boldsymbol{\theta}}_{t,0}^A$ | Initial prior of $\boldsymbol{\theta}^D$ for segment choice trials at $t$ | $\widehat{\boldsymbol{\Sigma}}_{t,0}^A$ | Initial prior of $\boldsymbol{\Sigma}^D$ for segment choice trials at $t$ |
| $\widehat{\boldsymbol{\theta}}_{t,m}^A$ | Updated prior of $\boldsymbol{\theta}^D$ for segment choice trial $m$ at $t$ | $\widehat{\boldsymbol{\Sigma}}_{t,m}^A$ | Updated prior of $\boldsymbol{\Sigma}^D$ for segment choice trial $m$ at $t$ |
| $\hat{x}_{i,j}$ | Observed $x_{i,j}$ | $\widehat{X}_{a^t}$ | Observed $X_{a^t}$ |
| $\tilde{\sigma}_{a^t}^{2,m}$ | Change in estimated variance of $\widehat{X}_{a^t}$ at $m$ | $\hat{\sigma}_W^2$ | Variance of observations |
| $\widetilde{\boldsymbol{\sigma}}(a^{t,m})$ | Change in $\widehat{\boldsymbol{\Sigma}}_{t,m}^A$ after observing $a^{t,m}$ | $W_{a^{t,m}}^{m+1}$ | Observation of the measurement of $a^{t,m}$ |
| $\widehat{\boldsymbol{\theta}}_{t,m}^A(a^{t,m})$ | Belief of the measurement of $a^{t,m}$ | $\boldsymbol{e}_{a^{t,m}}$ | Unit vector with 1 indicating $a^{t,m}$ |
| $\text{Var}^m$ | Variance after $m$ observations | $\varepsilon^{m+1}$ | Random error |
| $\kappa$ | Number of time steps in MAB | $n_{a^t}$ | Number of $a^t$ chosen |

| $\boldsymbol{\beta}^n$ | Vector of precision of $x_{i,j}$'s after $n$-th update | $\boldsymbol{\theta}^n$ | Vector of belief on mean of $x_{i,j}$'s after $n$-th update |
|---|---|---|---|
| $\boldsymbol{\beta}^W$ | Vector of precision of observations | $\boldsymbol{W}^{n+1}$ | Vector of observed flows |
| $\boldsymbol{\Omega}$ | Flow observation indicator matrix | $\omega$ | Vector indicating observations by 0 and 1 |
| $\sigma_\varepsilon^2$ | Observation error per flow | $L$ | Maximum route length |
| $P$ | Number of pilots | $L_P$ | Minimum pilot route length |
| $Q_P$ | Number of observations per pilot | $H$ | Diagonal matrix with $\sigma_{i,j}$'s |
| $B$ | Correlation matrix | | |

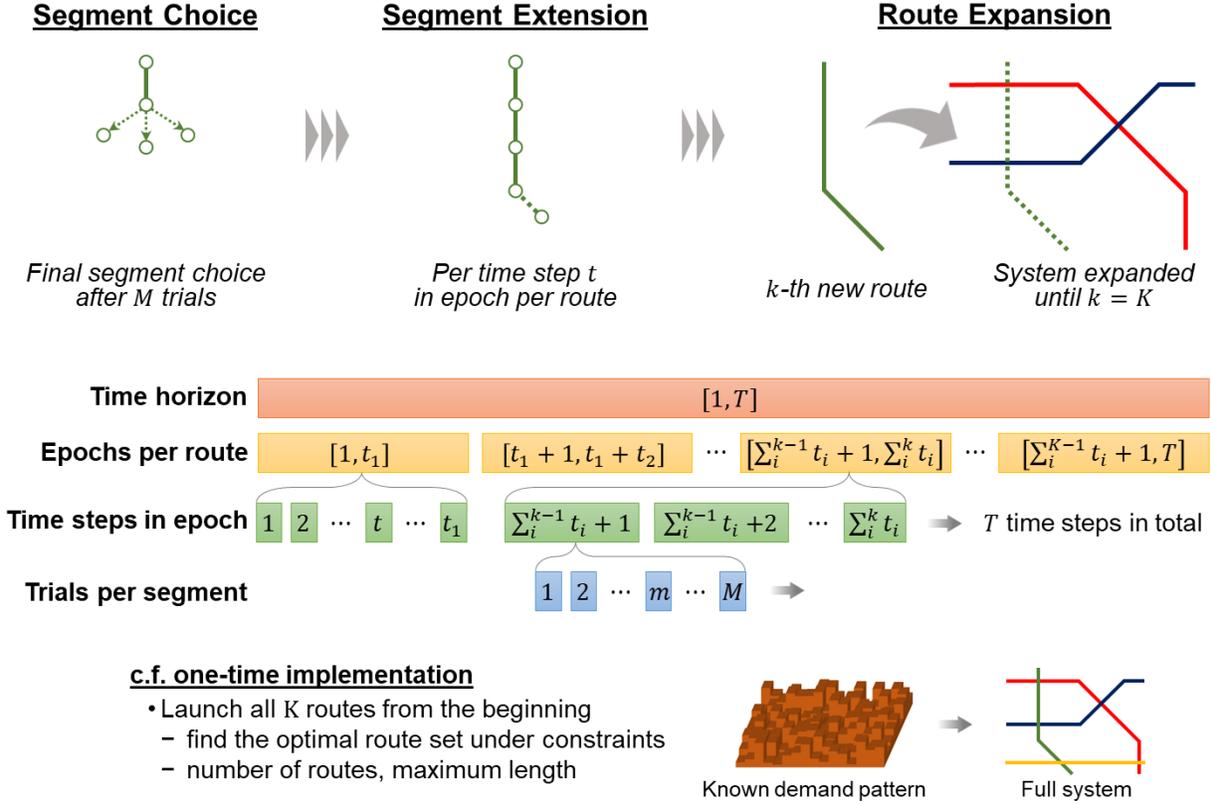

**Figure 2. Hierarchy of problem stage and time dimension.**

Suppose an operator plans $K$ candidate routes on a network $G(N,A)$ with nodes $n \in N$ and candidate bidirectional segments $a \in A$ over a time horizon $T$. A segment $a$ is denoted as $(p,q)$ to identify both ends. The operator decides to sequentially expand routes and apply a learning policy $\pi$ due to demand uncertainty. The time dimension is structured into a set of segment extension trials $M$ embedded within a segment extension time step $t$, consisting of a larger set of route expansion epochs in $T$. This is illustrated in **Figure 2**. For each $t$, there is a set of alternative segments $A^t$ that can be appended to the $k$-th route $R_k$ currently being extended and a set of corresponding nodes $N^t$. In a segment choice stage, one $a^t$ in $A^t$ is labeled as $a^{t,m}$ according to Definition 1. After $M$ trials, $R_k$ is extended to $a^{t,M}$ according to the evaluation, and the next time step $t+1$ is conducted. This repeats until the termination criteria are satisfied. In contrast, one-time implementation initiates all routes at the same time based on the knowledge about existing demand.

**Definition 1**. *The **sequential segment-level transit network design problem (SSTNDP)** selects a segment $a^t \in A^t$ across a sequential set of time steps $t \in T$ to maximize the cumulative value $\sum_{t \in T} v_{a^t}^{m,\pi}$, where $v_{a^t}^{m,\pi}$ is value of $a^t$ after $m$ trials under a learning policy $\pi$, and consists of a stochastic reward and an exploration term representing the potential benefit of exploration by collecting additional information from $a^t$, i.e. Eq. (5).*

$$a^{t,m} = \underset{a^t \in A^t}{\operatorname{argmax}} \, v_{a^t}^{m-1,\pi} \tag{5}$$

This study assumes that values of segments are dependent on OD demand, usually a unit by which passenger travel data are collected. For example, appending $a^{t,m}$ to $R_k$ can connect nodes of $R_k$ to $n^{t,m}$, the node corresponding to $a^{t,m}$. The broader coverage improvement is expected if the route intersects with others, enabling transfers among routes. Moreover, during the expansion, the operator should also consider correlations between different OD pairs of demand if they are assumed to be correlated. Definition 2 presents the information about OD demand used in the problem.

**Definition 2**. ***OD demand*** *of two nodes is the total bidirectional demand between them and considered as the size of a potential passenger group. True OD demand of node pair $(i,j)$ is denoted as $x_{i,j}$. $x_{i,j}$'s across all OD pairs are multivariate normal variates with $x \sim \mathbb{N}(\boldsymbol{\theta}^D, \boldsymbol{\Sigma}^D)$, where $\boldsymbol{\theta}^D$ is the array of $x_{i,j}$, and $\boldsymbol{\Sigma}^D$ is the true covariance matrix of $x_{i,j}$.*

The goal of the operator is to find the optimal sequence of segments appended and routes expanded, achieving the maximum cumulative value of the route system. In conventional transit network design, the segment expansion decisions require the assumption of a passenger assignment mechanism to load the flows onto the network. Typical mechanisms include the uncapacitated all-or-nothing assignment (Chakroborty and Wivedi, 2002; Cipriani et al., 2012; Iliopoulou and Kepaptsoglou, 2019), bilevel optimization with stochastic passenger route choice (e.g. Lam and Zhou, 2000) or hyperpath assignment (e.g. Verbas and Mahmassani, 2015). Because bilevel problems lead to nonconvexity that requires heuristics to solve, we rely on the simplest all-or-nothing assignment to focus on capturing the relationship between learning mechanisms and transit network structure with minimal noise in the numerical experiments.

Let $X^t$ be total OD demand coverage at $t$, the sum of $x_{i,j}$'s served by the current system. Demand coverage is the amount of demand accessible to the system designed through the framework. Mostly, it is equivalent to observed ridership but can also be prevailing demand for some occasions. If $\Delta X_{a^{t,M}}$ is the sum of additional $x_{i,j}$'s expected to be covered by appending $a^{t,M}$, $X^{t+1}$ after choosing $a^{t,M}$ is derivable as Eq. (6). Furthermore, this implies that $X^t$ consists of past values of $a^{t,M}$. This is possible since $\Delta X_{a^{t,M}}$'s for different $t$'s do not share common $x_{i,j}$ and are mutually exclusive.

$$X^{t+1} = X^t + \Delta X_{a^{t,M}} \tag{6}$$

$$= \sum_{i}^{t} \Delta X_{a^{t,M}}$$

A segment-level extension allows multiple $x_{i,j}$'s to use the system as their travel mode. Thus, when considering demand as a primary measure, the aggregation of node pair flows associated with segments is necessary. If $R_k$ with $b$ nodes attaches $a^t$ and extends to $n^t$, it can cover $b$ more node pairs. The number increases if there are routes intersecting with $R_k$. The framework does not simply add all OD flows but examines whether node pairs can be connected with the given route sets or not, per an uncapacitated all-or-nothing assignment. By defining $a^{t(q)}$ as the $q$-th segment of $A^t$ and $D^q$ as a set of possible $x_{i,j}$'s after integrating $a^{t(q)}$, Definition 3 explains the mean and covariance of options.

**Definition 3.** *For OD demand $x_{i,j}$ where $(i,j) \in D^q$, total **OD demand after choosing $a^{t(q)}$** is in Eq. (7).*

$$X_{a^{t(q)}} = \sum_{(i,j)}^{D^q} x_{i,j}. \qquad (7)$$

*The **covariance between $X_{a^{t(q_1)}}$ and $X_{a^{t(q_2)}}$** is shown in Eq. (8).*

$$\text{cov}(X_{a^{t(q_1)}}, X_{a^{t(q_2)}}) = \sum_{(i,j)}^{D^{q_1}} \sum_{(r,s)}^{D^{q_2}} \text{cov}(x_{i,j}, x_{r,s}). \qquad (8)$$

Both are elements of the mean vector $\boldsymbol{\theta}^A$ and the $|N^t| \times |N^t|$ covariance matrix $\boldsymbol{\Sigma}^A$, respectively. For example, Nodes 1, 2, and 3 in an existing route considers a new segment between Node 3 and 4. Then, $a^{t(q)} = (3,4)$ and $D^q = \{(1,2), (1,3), (2,3), (1,4), (2,4), (3,4)\}$. Thus, $X_{a^{t(q)}} = x_{1,2} + x_{1,3} + x_{2,3} + x_{1,4} + x_{2,4} + x_{3,4}$, while $\sigma^2_{a^{t(q)}}$ is the sum of covariances of 6×6 combinations of $x_{i,j}$. At $t+1$, however, $\boldsymbol{\theta}^A$ and $\boldsymbol{\Sigma}^A$ are renewed as $D^q$ has new elements added by expansion. Due to its dependence on $t$, they are labeled as $\boldsymbol{\theta}^A_t$ and $\boldsymbol{\Sigma}^A_t$.

Consequently, an online algorithm is needed to solve the SSTNDP, which requires a stochastic process for demand, a learning policy to translate a segment selection into a stochastic reward, and an optimization of the sequential segment selections to maximize the cumulative value. Section 3.2 presents the learning scheme needed to adopt optimal learning within sequential route expansion. Section 3.3 then presents the proposed algorithm for solving the SSTNDP.

### 3.2. Preliminaries: learning scheme
#### 3.2.1. Design of a prior

The learning policies depend on the set of priors being consistently updated to make their choices on segments. For the stochastic process, prior knowledge about the relationship between an option and its corresponding reward is essential for the choice and learning during sequential transit network design. In this problem, it represents the most recent information about OD demand in the given network that an operator can access. Since there are OD demand truths ($\boldsymbol{\theta}^D, \boldsymbol{\Sigma}^D$) and segment-level truths ($\boldsymbol{\theta}^A_t, \boldsymbol{\Sigma}^A_t$) not revealed, there should be four priors substituting each. First, $\widehat{\boldsymbol{\theta}}^D_0$ and $\widehat{\boldsymbol{\Sigma}}^D_0$ are initial priors of $\boldsymbol{\theta}^D, \boldsymbol{\Sigma}^D$ and updated to $\widehat{\boldsymbol{\theta}}^D_t$ and $\widehat{\boldsymbol{\Sigma}}^D_t$ at every $t$. Second, $\widehat{\boldsymbol{\theta}}^A_{t,0}$ and $\widehat{\boldsymbol{\Sigma}}^A_{t,0}$ in the segment level are prepared for the segment choice at $t$ and updated to $\widehat{\boldsymbol{\theta}}_{t,m}^A$ and $\widehat{\boldsymbol{\Sigma}}_{t,m}^A$ after each $m$.

Operators cannot directly observe these segment level priors, but they can be aggregated from OD demand priors by the same process with truths. They observe $\hat{x}_{ij}$, update corresponding elements in $\widehat{\boldsymbol{\theta}}_t^D$ and $\widehat{\boldsymbol{\Sigma}}_t^D$, and use them for synthesizing $\widehat{\boldsymbol{\theta}}_{t,m}^A$ and $\widehat{\boldsymbol{\Sigma}}_{t,m}^A$. The calculation is the same as the one in Definition 3, except for replacing the notation for $x_{ij}$ and $X_{a^{t(q)}}$ with $\hat{x}_{ij}$ and $\hat{X}_{a^{t(q)}}$.

If available, OD demand level priors can be recalled from external information sources or collected from pilot operations. Nevertheless, the initial pilots cannot cover all possible cases due to the lack of time and budget. Thus, a subset of options is observed and included in a prior. This may yield incomplete priors and cause empty elements in mean vectors or covariance matrices. Since they can disturb choices by distorting values of options, it requires appropriate assumptions to avoid incomplete priors. After the prior information is obtained, operators can designate the first segment to serve, and determine the direction of extension of the route.

### 3.2.2. Evaluation and choice

Because the calculation of $v_{a^t}^{m,\pi}$ is based on accumulated knowledge being continuously updated, $a^{t,m}$ from Eq. (5) can change during $M$ observations for a single extension. A current route is extended to $a^{t,M}$ at the final time step.

$v_{a^t}^{m,\pi}$'s of KG and KGCB are introduced in Literature Review, referring to Powell and Ryzhov (2012b). The computational complexity is lower with KG since it ignores correlations between flows. KG includes the calculation of: 1) $\tilde{\sigma}_{a^t}^{2,m}$, the change in estimated variance of $\hat{X}_{a^t}$ after the update by Eq. (9), 2) $\zeta_{a^t}^m$, the normalized influence of choosing $a^t$ by Eq. (10), and 3) $v_{a^t}^{m,KG}$, the knowledge gradient by Eq. (11). $\hat{\sigma}_W^2$ is the variance of observations.

$$\tilde{\sigma}_{a^t}^{2,m} = \hat{\sigma}_{a^t}^{2,m} - \hat{\sigma}_{a^t}^{2,m+1}$$
$$= \frac{\hat{\sigma}_{a^t}^{2,m}}{1 + \hat{\sigma}_W^2/\hat{\sigma}_{a^t}^{2,m}} \tag{9}$$

$$\zeta_{a^t}^m = -\left|\frac{\hat{X}_{a^t} - \max_{a^{t\prime} \neq a^t} \hat{X}_{a^{t\prime}}}{\tilde{\sigma}_{a^t}^m}\right| \tag{10}$$

$$v_{a^t}^{m,KG} = \tilde{\sigma}_{a^t}^m f(\zeta_{a^t}^m) \tag{11}$$

$$f(w) = w\Phi(w) + \phi(w) \tag{12}$$

Eq. (12) is a linear combination of $\phi(w)$, the standard normal density function, and $\Phi(w)$, the cumulative standard normal distribution shown in Eqs. (13) – (14).

$$\Phi(w) = \int_{-\infty}^{w} \phi(s)ds \tag{13}$$

$$\phi(w) = \frac{1}{\sqrt{2\pi}} e^{-\frac{w^2}{2}} \tag{14}$$

On the other hand, KGCB requires a more complex derivation. Recalling Eq. (4), $\widetilde{\sigma}(a^{t,m})$ and $Z^{m+1}$ should be defined to derive the expected value of Eq. (15) for all option $a^t$s. Eq. (16) indicates the change in $\widehat{\Sigma}^A_{t,m}$ after measuring option $a^{t,m}$. $Z^{m+1}$ in Eq. (17) is a random variable in the form of a standard normalization of the difference between the observation $W^{m+1}_{a^{t,m}}$ and $\widehat{\boldsymbol{\theta}}^A_{t,m}(a^{t,m})$, the belief of the measurement of $a^{t,m}$. $W^{m+1}_{a^{t,m}}$ is also a random variable due to the uncertainty in observations. Eq. (18) shows the equivalence between the denominator in Eqs. (16) – (17).

$$\widehat{\boldsymbol{\theta}}^A_{t,m+1} = \widehat{\boldsymbol{\theta}}^A_{t,m} + \widetilde{\boldsymbol{\sigma}}(a^{t,m}) Z^{m+1} \tag{15}$$

$$\widetilde{\boldsymbol{\sigma}}(a^{t,m}) = \frac{\widehat{\Sigma}^A_{t,m} \boldsymbol{e}_{a^{t,m}}}{\sqrt{\widehat{\Sigma}^A_{t,m}(a^{t,m}, a^{t,m}) + 1/\widehat{\sigma}^2_W}} \tag{16}$$

$$Z^{m+1} = \frac{W^{m+1}_{a^{t,m}} - \widehat{\boldsymbol{\theta}}^A_{t,m}(a^{t,m})}{\sqrt{\mathrm{Var}^m \left( W^{m+1}_{a^{t,m}} - \widehat{\boldsymbol{\theta}}^A_{t,m}(a^{t,m}) \right)}} \tag{17}$$

$$\begin{aligned} \mathrm{Var}^m \left( W^{m+1}_{a^{t,m}} - \widehat{\boldsymbol{\theta}}^A_{t,m}(a^{t,m}) \right) &= \mathrm{Var}^m \left( \widehat{\boldsymbol{\theta}}^A_{t,m}(a^{t,m}) + \varepsilon^{m+1} \right) \\ &= \widehat{\Sigma}^A_{t,m}(a^{t,m}, a^{t,m}) + 1/\widehat{\sigma}^2_W \end{aligned} \tag{18}$$

Eq. (15) is now a linear function with the independent variable $Z^{m+1}$, and Eq. (4) seeks the expected values of maximum $\widehat{\boldsymbol{\theta}}^A_{t,m}(a^t)$ for all options. If a set of lines are sorted by $\widetilde{\boldsymbol{\sigma}}(a^t)$ in ascending order, $c_j$s, the intersections of lines are identified. A line that dominates others on $[c_j, c_{j+1}]$ represents the maximum $\widehat{\boldsymbol{\theta}}^A_{t,m}(a^t)$ within the section. After eliminating curves without sections dominating others, the number of remaining curves is $J$, and $\widetilde{\sigma}_j(a^t)$ is defined as the slope of $j$-th curve with $a^t$. Since some lines may have no section to be better than others, intersections are labeled with different subscripts. After ordering curves, it is possible to estimate Eq. (4) with Eq. (19).

$$v^{m,KGCB}_{a^t} = \sum_{j=1}^{J} \left( \widetilde{\boldsymbol{\sigma}}_{j+1}(a^t) - \widetilde{\boldsymbol{\sigma}}_j(a^t) \right) f(-|c_j|) \tag{19}$$

where $f$ is the function in Eq. (12). Details of this derivation can be found in Frazier et al. (2009) and Powell and Ryzhov (2012b).

$v^{m,MAB}_{a^t}$ is in Eq. (20). The stochastic factor is explained with an upper confidence bound which can achieve logarithmic regret as $k$ increases (Auer et al., 2002; Lattimore, 2016).

$$v_{a^t}^{m,MAB} = \hat{X}_{a^t} + \sqrt{\frac{2\log \kappa}{n_{a^t}}} \tag{20}$$

where $\kappa$ is the number of time steps, and $n_{a^t}$ is the number of $a^t$ chosen during the evaluation. More observations of a certain $a^t$ lead to smaller disturbance of its $v_{a^t}^{m,MAB}$, increasing the predictability of $\hat{X}_{a^t}$.

### 3.2.3. Observation and knowledge update

An option returns a reward when chosen. Because the operator wants to attract as many customers as possible, the covered demand is considered the reward measure. Under uncertainty, expected rewards need to be estimated and the same action applied twice may not cause the same result each time.

The model should update the archived knowledge from the observation of the choice result. For route design, an expanded system can cover more passengers, causing changes to the previous knowledge about demand patterns. Consequently, the system should update the knowledge after each expansion until the final route is implemented.

There are three types of knowledge that should be updated: prior mean, prior variation, and prior covariance matrix. For an easier update, a precision $\beta$, the inverse of variance, replaces $\sigma^2$ ($\beta = 1/\sigma^2$). It captures how accurate values can be observed. If only some flows are assumed to be correlated, priors should be separated accordingly. First, uncorrelated flows are considered random variables following independent normal distributions. Then the update is done with a simple Bayesian prior update procedure like Eqs. (21) – (22) (Powell and Ryzhov, 2012).

$$\theta^{n+1} = \frac{\beta^n \theta^n + \beta^W W^{n+1}}{\beta^n + \beta^W} \tag{21}$$

$$\beta^{n+1} = \beta^n + \beta^W \tag{22}$$

where $\boldsymbol{\theta}^n$ and $\boldsymbol{\beta}^n$ are current beliefs on means and precisions of flows, $\boldsymbol{\beta}^W$ is the precision of observations, $\boldsymbol{W}^{n+1}$ is the observed flow, and $\boldsymbol{\theta}^{n+1}$ and $\boldsymbol{\beta}^{n+1}$ are updated beliefs.

Correlated flows require more complicated update procedures because of the incomplete observations. Since any observation cannot be complete unless the operating route system covers all flows, the system can only collect information on a subset of flows. Additional consideration of omitted flows is essential for correlated flows by introducing a matrix $\boldsymbol{\Omega}$ indicating whether a flow is observed on its diagonal. Eqs. (23) – (25) are expressions for updating correlated priors (Perkerson, 2020).

$$\boldsymbol{\Omega} = \text{Diag}(\omega/\sigma_\varepsilon^2) \tag{23}$$

$$\boldsymbol{\theta}^{n+1} = \left(\boldsymbol{\Omega} + \boldsymbol{\Sigma}^{n-1}\right)^{-1} \left(\boldsymbol{W}^{n+1^T}\boldsymbol{\Omega} + \boldsymbol{\theta}^{n^T}\boldsymbol{\Sigma}^{n-1}\right)^T \tag{24}$$

$$\boldsymbol{\Sigma}^{n+1} = \left(\boldsymbol{\Omega} + \boldsymbol{\Sigma}^{n-1}\right)^{-1} \tag{25}$$

### 3.3. Proposed AI-based sequential segment-level transit network design algorithm

There are differences between route design and conventional optimal learning. First, route design does not repeat evaluation on the same option set. After a route is extended by a link, evaluation of the next extension is based on a different option set. Next, the number of observations and operations are relatively limited compared to other optimal learning applications. For instance, if a system aims for the number of clicks on online news articles, it can collect responses of readers due to its ubiquity. However, lower operational flexibility of mobility services prohibits operators from conducting massive experiments. Moreover, rewards regarding available options during the evaluation are aggregated OD flows, requiring the aggregation for every extension. Considering these differences, the following new system design algorithm is proposed.

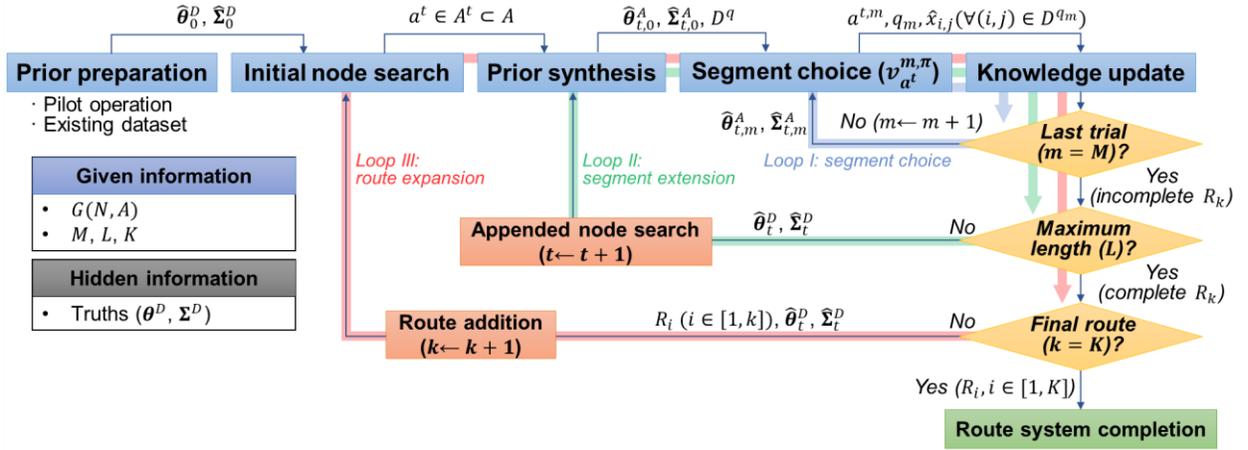

**Figure 3. Proposed learning-based sequential segment-level network design algorithm.**

**Figure 3** describes an algorithm of the AI-based segment-based extension using optimal learning. In this algorithm, the operator should prepare $\widehat{\boldsymbol{\theta}}_0^D$ and $\widehat{\boldsymbol{\Sigma}}_0^D$ from available sources such as pilot service operations or existing dataset regarding OD demand since the truths $\boldsymbol{\theta}^D$ and $\boldsymbol{\Sigma}^D$ are not revealed. Once the initial terminal (or both ends after the first segment choice) of $R_k$ is designated, all segments that can be attached to the current $R_k$ become options. $\widehat{\boldsymbol{\theta}}_0^D$ and $\widehat{\boldsymbol{\Sigma}}_0^D$ are aggregated to the segment level using transit assignment to form $\widehat{\boldsymbol{\theta}}_{t,0}^A$ and $\widehat{\boldsymbol{\Sigma}}_{t,0}^A$ used for the segment evaluation based on $\pi$. The segment with the highest $v_{a_t^t}^{m,\pi}$ is labeled as the optimal option $a^{t,m}$ at the $m$-th trial, and its associated $\hat{x}_{i,j}$ are observed. After $M$ trials are observed through Loop I, the operator moves on to the next extension and repeats Loop II until $R_k$ is complete. Consequently, $k$ times of Loop III yields the complete route system that the operator desired.

This study considers three policies as candidates of the learning policy in this proposed AI-based network design algorithm, where they share common aspects in the algorithm including data processing and input, feedback of interim results, and output production. The evaluation part of the proposed algorithm can be replaced if someone needs another learning policy. **Algorithm 1** explains the flow of the proposed methodology to solve the model in Eq. (5).

**Algorithm 1. Proposed AI-based sequential segment-level transit network design algorithm**

1. Prepare network $G(N, A)$ and node pair flows $\{x_{i,j} | i, j \in N\}$
2. Input simulation settings: number of routes required $K$, maximum route length $L$, number of pilots $P$, minimum pilot route length $L_P$, number of observations

|   |   |
|---|---|
|   | per pilot $Q_p$, number of observations per extension $M$, learning policy $\pi$ |
| 2.1. | If available, import priors from existing knowledge ($\hat{\boldsymbol{\theta}}_{a^0}^D, \hat{\boldsymbol{\Sigma}}_{a^0}^D$). |
| 3. Conduct pilots | |
| **For** $p = 1$ to $P$ **do** | |
| 3.1. | Randomly choose two nodes $(i,j)$ as terminals. |
| 3.2. | Operate a route between $(i,j)$, longer than $L_P$. |
| 3.3. | Observe $Q_p$ operations on chosen route. |
| 3.4. | From observed $x_{i,j}$s, create $\hat{\boldsymbol{\theta}}_{a^0}^D$ and $\hat{\boldsymbol{\Sigma}}_{a^0}^D$ if not existent. Otherwise, update them. |
| **End For** | |
| 4. Implement segment-level extension | |
| **For** $k = 1$ to $K$ **do** | |
| 4.1. | Designate an initial node as a starting terminal of $R_k$. |
| **While** $|R_k| < L$ **do** | |
| 4.2. | Identify available links $a^t \in A^t$ from both ends of $R_k$. |
| 4.3. | Estimate $v_{a^t}^{m,\pi}$ of links. |
| **For** $q = 1$ to $|A^t|$ **do** | |
| 4.3.1. | Identify $x_{i,j}$'s covered by the route system with $a^{t(q)}$ appended. |
| **End For** | |
| 4.3.2. | Aggregate $\hat{\boldsymbol{\theta}}_{a^0}^D$ and $\hat{\boldsymbol{\Sigma}}_{a^0}^D$ of $x_{i,j}$'s to segment-level ($\hat{\boldsymbol{\theta}}_{a^t}^A, \hat{\boldsymbol{\Sigma}}_{a^t}^A$) using transit assignment. |
| **For** $m = 1$ to $M$ **do** | |
| 4.3.3. | Input priors to optimal learning policy and find which $a^{t(q)}$ to observe. |
| 4.3.4. | Observe $x_{i,j}$'s corresponding to $a^{t,m}$ and updated knowledge. |
| **End For** | |
| 4.4. | Append $a^{t,M}$ to $R_k$ according to Eq. (5). |
| **End While** | |
| **End For** | |
| 5. Yield route system and performance measures | |

**Algorithm 1** is coded in MATLAB and is made available on Zenodo (Yoon, 2023). Note that it makes use of KGCB and KG codes provided by Castle Labs (2022a,b). The impact of the network size is limited in **Algorithm 1**; it only affects the designation of the initial starting terminal. Instead, the complexity of this algorithm is determined by $K$, $L$, $M$, and mean $|A^t|$, which have low correlation with the network size. The more important issue regarding scalability is in terms of the time frame and algorithm loops. Planners need to consider trade-offs between computational efficiency and reliability of accumulated data. One example timeframe can be expanding a system every few weeks and completing routes every few months, which would be more computationally costly and reliable than a system that considers expansions every few months and completes routes every few years. Detailed numbers will vary regarding parameters.

Different learning policies also have varying computational complexities that affect the speed of algorithms. For example, KGCB requires matrix calculation consuming much time compared to the others. However, in practice, this cost difference between learning policies may have limited impact on performance due to the long timeframe compared to the computation time. For example, a service operator cannot operate different routes day by day due to the confusion that users and drivers would experience. Instead, they may operate the same route for several days or weeks to observe the reward – the observed demand. This means that the operator should have sufficient time to determine which segment to append in the next period.

## 4. Numerical Experiment

The purpose of the numerical experiment is to validate whether the proposed algorithm can establish a reasonable route system and evaluate its performance under three different learning policies. This is accomplished using two numerical experiments. First, a small grid network and travel demand within the region are generated. Second, we create an experimental network based on New York City Public Use Microdata Areas (NYC PUMAs). The result from RHTS is assumed as the source of an initial prior.

### 4.1. Simple Grid Networks
#### *4.1.1. Problem and Network Illustrations*

For the demonstration of the segment-level system expansion, randomly generated trips travel on a 5-by-5 grid network with bidirectional links. In this region, an operator seeks a route set that serves the demand the best, but the knowledge about trips is limited to the result of pilot services. A set of 50 simulations are conducted in this network to derive average performances. **Figures 4**(a) and **4**(b) represent true flows and observations from pilot services, respectively. The thickness of a line is proportional to the amount of flow. This shows an example of a knowledge gap between the truth and observed information. While the true flow pattern remains constant during simulations, observed flows vary as routes for pilot services are randomly generated in each of the 50 runs.

Several conditions are given in this experiment: 1) the final system consists of three routes with four nodes, 2) five pilot services are operated for 10 time periods, 3) 198 node pair flows are assumed to be correlated among 300 available flows, 4) transfer between routes are allowed for once, and 5) standard deviation $\sigma_{i,j}$ is assumed to be 10-30% of $x_{i,j}$.

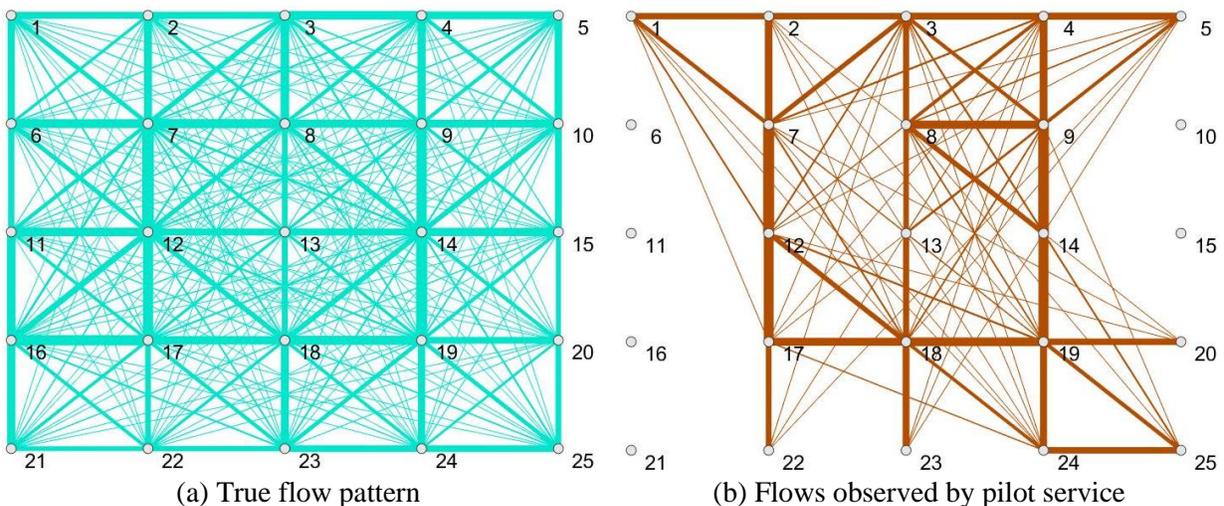

(a) True flow pattern  (b) Flows observed by pilot service
**Figure 4. Example of knowledge gap between truth and observation to illustrate the incompleteness of observations from initial pilots.**

*4.1.2. Experiment Inputs*

Two kinds of demand information are required in the experiment: truth and prior. All truths of mean, standard deviation, and covariance are assumed, and priors are derived from simulated pilot services conducted in the experiment.

First, flows between node $i$ and $j$ are assumed to be proportional to the multiplication of trip production constant of $i$ and trip attraction constant of $j$ and the inverse of square of distance. When flows calculated, true standard deviations are generated to lie between 10% and 30% of the true mean. The true covariance matrix $\boldsymbol{\Sigma}^D$ is derived from the multiplication of matrices shown in Eq. (26).

$$\boldsymbol{\Sigma}^D = \boldsymbol{HBH} \qquad (26)$$

where $\boldsymbol{H}$ is a diagonal matrix with $\sigma_{i,j}$ on its diagonal, and $\boldsymbol{B}$ is the correlation matrix generated by the constant correlation model (Hardin et al., 2013), satisfying the positive semidefinite (PSD) condition.

Second, priors are derived from observations during pilot services. Due to the incompleteness of pilots only partially covering the region, observations are made with subsets of flows. If a flow is observed, its prior mean and standard deviation can be calculated. Otherwise, both are set to zero, indicating the absence of observations. In the same manner, zeros are assigned to covariance matrix elements of omitted observations. Nonetheless, diagonal elements should not be zeros to maintain the PSD condition of the covariance matrix. Alternatively, we replace zeros with 1% of the mean. For replication purposes, the truth and prior data for this test instance can both be found at https://zenodo.org/badge/latestdoi/636483323.

*4.1.3. Results*

**Figure 5** explains the generated route sets that vary as different policies are applied in one sample run. A link is overlapped in the result of MAB and KG because both choose based on the remaining unserved flows of corresponding nodes calculated from priors, regardless of the current service availability. While outputs of MAB and KG remain similar except for the last segment extension of the 3rd route, 3-3, the 1st route of KGCB initially takes another direction, resulting in the avoidance of overlapping links. Nevertheless, its system shares many common points with two previous results, and it implies a similarity of evaluation processes of policies, depending on given priors, especially for the mean.

The demand coverage rate, a percentage of covered demand compared with total demand, is the highest when using KGCB on average. According to the result, KGCB shows about a 2pp increase in the demand coverage rate over the other two mechanisms. Moreover, the standard deviation of KGCB is the lowest, implying the stability of the result. This experiment is primarily to provide a replicable set of instances testing. A sequential route expansion result can be influenced by the initial network condition and pilot service plan. To improve the reliability of the comparison, a more realistic case is investigated in the next experiment.

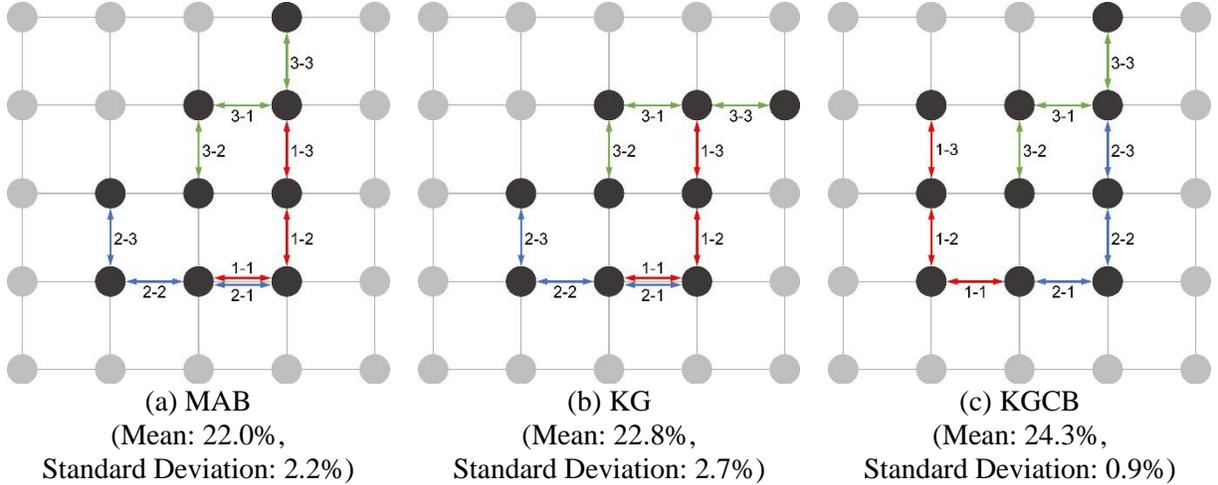

| (a) MAB | (b) KG | (c) KGCB |
| --- | --- | --- |
| (Mean: 22.0%, | (Mean: 22.8%, | (Mean: 24.3%, |
| Standard Deviation: 2.2%) | Standard Deviation: 2.7%) | Standard Deviation: 0.9%) |

**Figure 5. Underlying processes to route set outcome from policies with demand coverage rate.**

### 4.2. NYC PUMA-based Network

#### 4.2.1. Problem and Network Illustrations

The operator plans for sequential implementation of 5 routes with 8 nodes each. During segment-level extensions, they can intersect with others to allow one-time transfer for riders to enhance the coverage of the system. 30 pilot routes are planned, and each can observe flows 10 times. The minimum length of pilot routes is 6 nodes. The objective value is demand coverage. To make the problem simple to focus the evaluation on learning framework comparison with minimal noise, travel time considerations by riders, congestion effects on links, vehicle capacity limitations are all assumed to be negligible.

The generated network consists of 55 nodes and 123 bidirectional links as shown in **Figure 6**. For simplicity, it is assumed that link costs for system implementation are equal to make all options require the same cost during the segment extension. Since the number of possible node pairs is 1,485 (=55×54/2), the prior mean vector should have 1,485 elements, and the size of the covariance matrix becomes 1,485-by-1,485 if correlations between all flows exist. Among 1,485 OD flows, 300 are assumed correlated. Therefore, priors for both correlated and uncorrelated flows should be updated separately.

There are five boroughs in NYC; Manhattan, Brooklyn, Staten Island, Queens, and Bronx. First, we assumed flows may be correlated with each other if they connect the same pair of boroughs. In addition, Manhattan and Bronx are integrated into one "megaborough" due to their high connectivity. With four zones, there are 10 possible types of flows including inter- and intraborough connections. However, 7 remain after flows connected to Staten Island are aggregated into one group to reflect being an isolated island. **Figure 7** is the OD flow matrix of RHTS and overlapped groups. In this experiment, they are named "clusters."

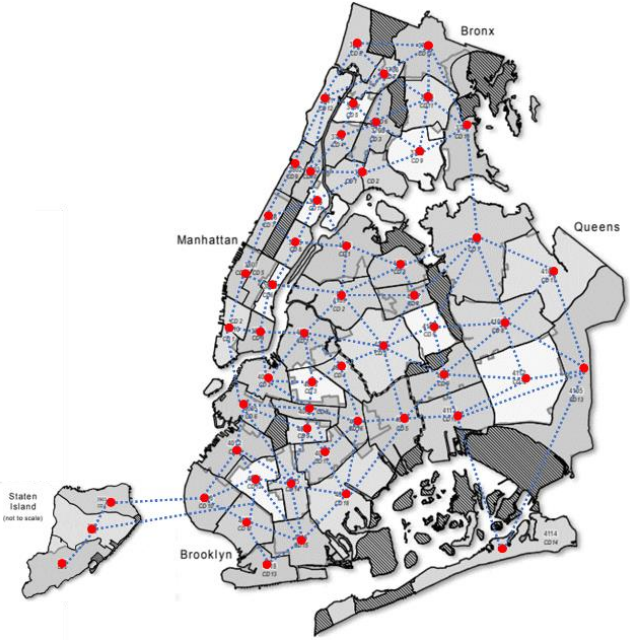

**Figure 6. Experimental network with NYC PUMA.**

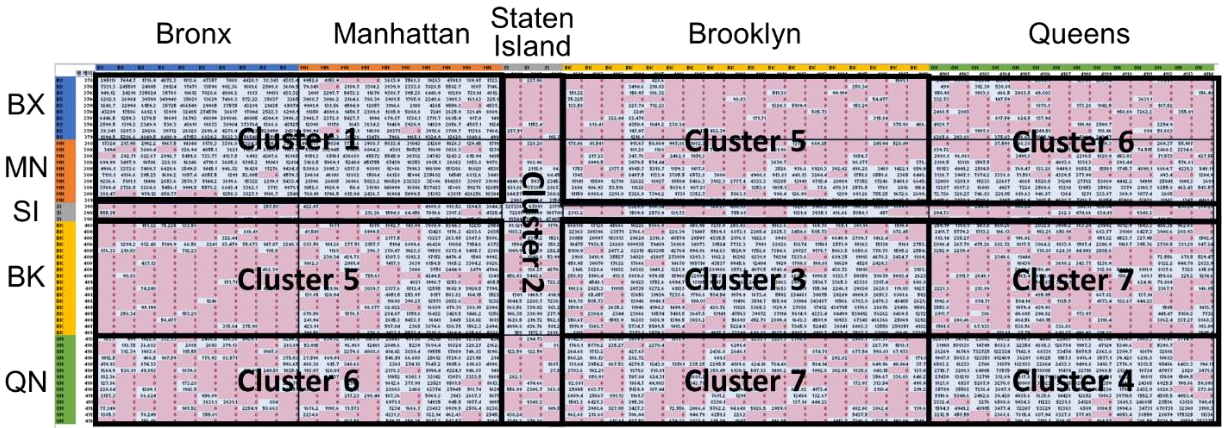

**Figure 7. OD flow matrix of RHTS and clusters of correlated flows.**

Any node can be a route terminal when ignoring geography. Nevertheless, since the methodology focuses on demand coverage, nodes located on the periphery are less likely to be chosen due to its disadvantage of limited direction of extension. Instead, nodes in the central area have higher chances to become an initial terminal.

*4.2.2. Simulation inputs*

As shown in the previous experiment, the truth and prior need to be defined. We use the RHTS data as the initial prior. After assuming priors, synthetic truths are reverse-engineered from them. As a result, truths in this experiment are artificially constructed under these assumptions.

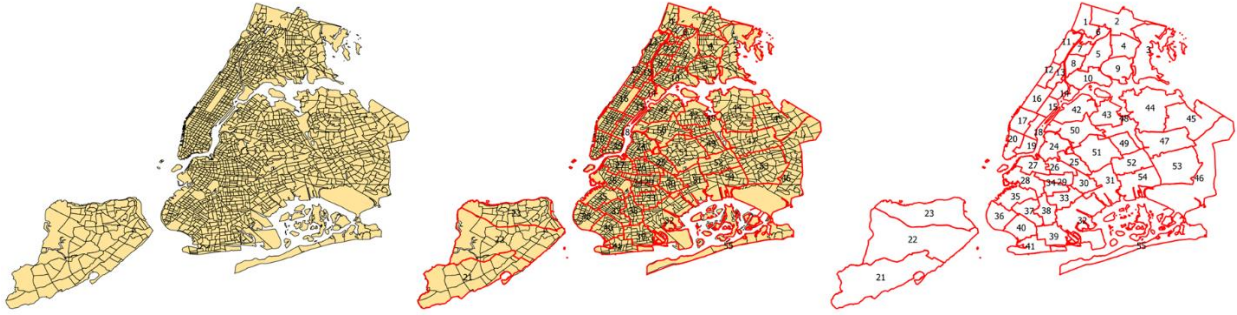

**Figure 8. Aggregation of 1622 Census tracts (left) to 55 PUMAs (right) by overlapping (middle).**

First, the flow information from the RHTS becomes a prior mean. As the survey was conducted in Census tract level, the result is aggregated to PUMA level as shown in **Figure 8**. Meanwhile, the RHTS is a single observation, and other priors such as variances and covariances are impossible to derive. Thus, prior standard deviations of flows are initially simply assumed, for example, 5% of means while covariances are derived from the result of pilot services as the previous experiment does.

Second, true standard deviations are selected from assumed ranges. For instance, a flow of 10,000 units may vary with a standard deviation between 5% and 19% of 10,000. The range changes as the amount of flow varies. In addition, scenarios are prepared for three different levels of flow variation: Low, Middle, and High. The variation of OD flow means the fluctuation of observed demand for each time period. The higher variation means the higher uncertainty that operators would face. While the lower limit is equivalent, different upper limits are set for each level. With the generated true standard deviations and prior means, a true mean vector can be chosen from random numbers from the inverse normal distribution while true covariance matrix is calculated from the true standard deviations. The truth and prior data for this case study are provided in https://zenodo.org/badge/latestdoi/636483323 for replication purposes.

### *4.2.3. Results*

**Figure 9** shows cumulative curves of covered demand in 15 scenarios (5 different demand patterns and 3 flow variation levels). Every scenario is simulated 100 times and averaged results are reported. The learning policies are compared with other benchmarks for validation.

First, yellow transparent curves result from a greedy policy that chooses the option with the highest expected value without considering exploration.

Second, blue transparent curves represent the CR reference policy (Chow and Sayarshad, 2016) that estimates the distribution of maximum value policy by assuming that randomly selected policies are samples of a Weibull-distributed policy. The advantage of using this reference policy over a purely oracle or competitive ratio (Karp, 1992) is that the policy is consistently defined to represent the best value that can be obtained with a committed design without any online reoptimization, one that is sensitive to both the underlying network characteristics and the stochasticity of the uncertain environment. For example, for the same instance over a time horizon, changing the underlying variance of a stochastic variable may yield the same average competitive ratio when solved over multiple simulated runs since the average of the best performance would not change. However, the CR policy would result in a different cumulative distribution for the maximum value obtainable with the best committed design. This means that while changing scenarios may make it hard to compare outputs with each other directly, they can be compared relative to the consistently defined CR policy. Plotted curves illustrate differences among policies in terms of their performance.

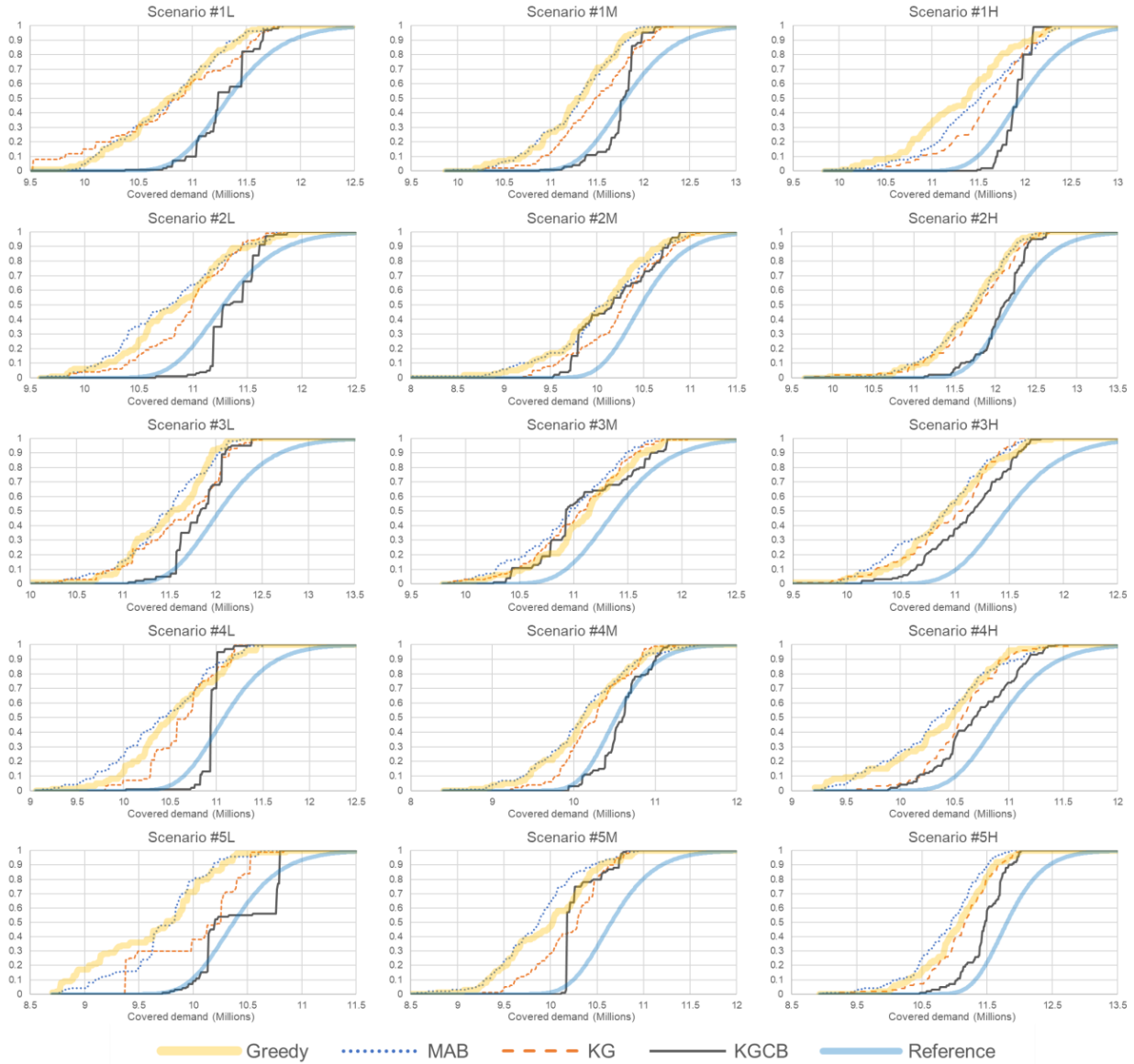

**Figure 9. Cumulative curves of covered demand from different learning policies.**

In most diagrams, the dark gray curve representing cumulative covered demand of KGCB tends to be located at the right of the curves from the other two policies.

**Figure 10** is the plot indicating gap between performance of 50-percentiles of the learning policies and the CR reference policy curves in **Figure 9**. There is no firm consistency observed in the order of performance, and incorporating learning in system design even deteriorates the performance in some scenarios. However, in some cases, KGCB even outperforms the 50-percentile CR policy (though they never outperform the 100-percentile). In general, KGCB yields the best output relative to the reference policy, and its ability to consider correlations among demand may support this advance.

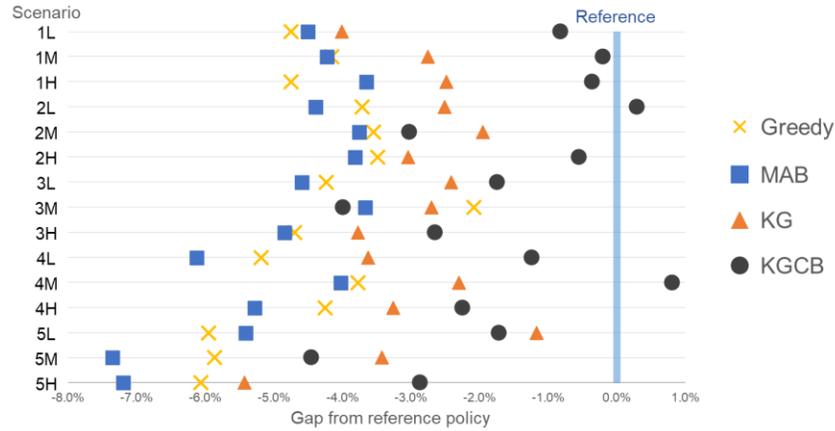

**Figure 10. Gap between learning and reference policy at 50-percentile.**

**Remark 1.** *KGCB outperforms in at least a portion of the 100 simulations through all 15 NYC scenarios, in some cases outperforming the 50-percentile CR policy, which suggests there is value in capturing correlated beliefs in exploration when solving the SSTNDP.*

Detailed numerical results are reported in **TABLE 2**. $t$-tests are used to verify if differences in distributions of covered demand from learning policies are statistically significant. From observation, there are instances where MAB and KG underperform even the greedy policy, and none outperform any portion of the CR policy. Meanwhile, KGCB does outperform some percentiles of the CR policy in some scenarios. While comparing MAB with KG is not significant in 5 scenarios, the number of statistically significant differences is 14 for MAB-KGCB and 13 for KG-KGCB, out of the 15 scenarios. This means that better performance of KGCB is not only graphically shown but also statistically proven.

**Remark 2.** *KGCB appears to perform better in the NYC scenarios when the demand variances are lower.*

It is expected that gaps between KGCB and other policies may become larger when working with a network with highly variable flows. However, a statistically significant trend between objective values and flow variations is not found, implying that the performances of learning policies in our framework are not significantly impacted by the flow variations. Instead, the demand pattern appears to have more influence on the result.

**TABLE 3 Descriptive Statistics of Simulated Covered Demand of Policies and $t$-test Results**

| Scenario | Descriptive statistics (mean & std.dev in million) | | | Two-tail $t$-test result ($p$-value) | | |
|---|---|---|---|---|---|---|
| | MAB | KG | KGCB | MAB vs KG | MAB vs KGCB | KG vs KGCB |
| #1L | 10.78 (0.49) | 10.80 (0.64) | 11.29 (0.26) | *0.854* | *** | *** |
| #1M | 11.27 (0.45) | 11.48 (0.43) | 11.76 (0.22) | 0.001 | *** | *** |
| #1H | 11.48 (0.55) | 11.59 (0.45) | 11.90 (0.14) | *0.133* | *** | *** |
| #2L | 10.78 (0.52) | 10.95 (0.42) | 11.37 (0.21) | 0.001 | *** | *** |

| #2M | 10.05 (0.58) | 10.24 (0.46) | 10.18 (0.41) | 0.011 | _0.057_ | _0.376_ |
|---|---|---|---|---|---|---|
| #2H | 11.68 (0.48) | 11.75 (0.52) | 12.08 (0.31) | _0.314_ | *** | *** |
| #3L | 11.46 (0.46) | 11.61 (0.52) | 11.83 (0.27) | 0.024 | *** | 0.000 |
| #3M | 10.94 (0.46) | 11.04 (0.44) | 11.10 (0.45) | _0.151_ | 0.016 | _0.318_ |
| #3H | 10.85 (0.46) | 10.93 (0.44) | 11.12 (0.38) | _0.209_ | *** | 0.001 |
| #4L | 10.41 (0.54) | 10.64 (0.35) | 10.95 (0.12) | 0.000 | *** | *** |
| #4M | 10.10 (0.58) | 10.25 (0.41) | 10.59 (0.29) | 0.028 | *** | *** |
| #4H | 10.35 (0.52) | 10.57 (0.34) | 10.70 (0.37) | 0.001 | *** | 0.007 |
| #5L | 9.73 (0.40) | 10.03 (0.45) | 10.40 (0.35) | *** | *** | *** |
| #5M | 9.85 (0.46) | 10.20 (0.37) | 10.30 (0.20) | *** | *** | 0.022 |
| #5H | 10.87 (0.55) | 11.12 (0.50) | 11.46 (0.33) | 0.001 | *** | *** |

Note: Underlined and italic numbers mean statistically insignificant $p$-values with $\alpha = 0.05$. *** are ones with very strong significance.

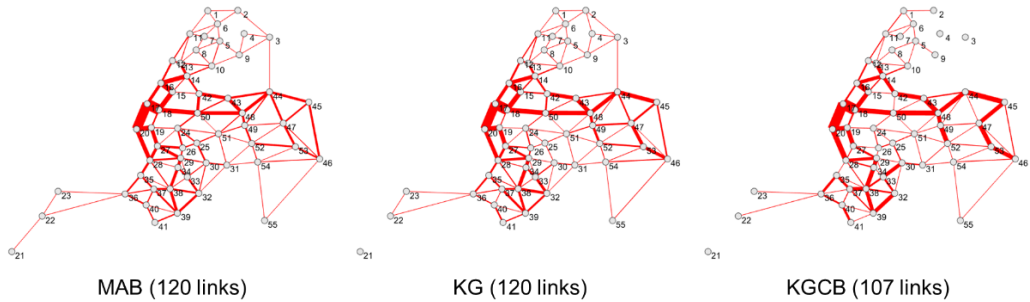

**Figure 11. Link choice frequencies of learning policies (based on Scenario #5H) averaged over 100 simulations each.**

**Figure 11** reports the networks with links weighted proportionately to the number of choices made during the 100 simulations under the three different learning policies for the proposed network design algorithm to highlight the resulting differences. Most parts of the networks are similar, but KGCB tends to focus more on the Queens neighborhoods and less on the Bronx areas. The result of chi-squared tests between policies suggests the differences in their link choice behaviors are statistically significant ( $\chi^2_{MAB-KG} = 187.53\ (d.f. = 120)$ , $\chi^2_{MAB-KGCB} = 675.20\ (d.f. = 119)$, and $\chi^2_{KG-KGCB} = 498.14\ (d.f. = 119)$ ). The main reason can be the different way that they evaluate links and the extent of the exploration. Since KGCB uses more information than MAB and KG, it decreases the chance of exploring unknowns.

## 5. Conclusion

Considering the limitation of information about prevailing demand and its variability, designing a mobility route system with one-time planning by assuming static demand pattern may yield lower system performance. This is especially the case for transit networks built off emerging technology with limited user experience: automated vehicle fleets, electric buses, microtransit, semi-flexible routes, first-last mile hubs, etc. As an alternative, this study presents a new type of transit network design problem based on using information in a sequential manner, the SSTNDP. We further propose a new algorithm for extending existing routes and expanding new routes, making use of optimal learning techniques to consider the difference between the existing knowledge and actual data observed. The proposed approach sequentially extends the system by choosing an option within a given set based on the most recent knowledge updated after observing the consequence of choices made, as an optimal learning system.

Three alternative learning policies, KGCB, KG, and MAB are integrated into the proposed framework and compared in the numerical experiment with a grid network and artificial NYC PUMA network. Among two case studies with different truths and flow variation levels, the proposed algorithm with integrated KGCB tends to show the best performance represented by mean covering demand. Since it is the only policy that considers correlations between flows, it may have a stronger ability to explore other options.

One of the biggest advantages of this methodology is the applicability to the region without good quality demand pattern data. It reduces the dependency on existing data sources and promotes direct information accumulation which makes the system more responsive to gaps that could not be captured by previous data collection. Furthermore, due to its gradual expansion of the system, it can be more attractive when the project budget is divided over time and sequentially executed.

For simplicity, this approach excludes some details that exist in the real world, such as no limitations to the number of transfers between routes and actual segment extension costs, and the use of simple all-or-nothing assignment rules or simplifications that may not completely capture a specific transit service technology like bus, rail, or on-demand microtransit without further adaptation. Without transfer, a designed route system is equivalent to a set of individual routes that passengers cannot have benefits from the route expansion. The main difference should be the reward of choices. Similarly, more realistic assignment mechanisms can be assumed when implementing this framework in practice, although heuristics will need to be used and greater noise will be added to the results. While it only includes demand within the newly extending route in this study, it is necessary to consider additional OD flows connected to other routes accessible from the current route. Moreover, since the approach refers to optimal learning policies that assume that costs of choosing different options are identical, extending routes to various links in this methodology also requires the same cost.

Thus, anticipated future research may concentrate on the relaxation of these assumptions to improve the applicability of the framework in the real world. This framework opens many new avenues of future research in sequential transit network design, such as better design of priors using initial surveys or pilots, alternative network structures like zone-based service provision, integration of bus performance with traffic congestion, or networks that can grow and contract over time. An alternative set can include not only attaching adjacent segments to the current route but also initiating a new route, providing more options to planners. Furthermore, the framework can conduct a multifaceted evaluation on expected benefit and cost associated with each choice so that planners can consider both the performance and financial feasibility simultaneously. Lastly, there can be an effort to apply the sequential network design to an existing transit network to modify the system to corresponding demand patterns as a variation of evolutional model, perhaps as a digital twin implementation.

## ACKNOWLEDGEMENTS

This study is supported by C2SMART, a USDOT Tier 1 University Transportation Center, and NSF CMMI-1652735. The work forms one chapter of the dissertation submitted to the Faculty of the New York University Tandon School of Engineering in partial fulfillment of the requirements for the degree Doctor of Philosophy (Transportation Planning & Engineering), January 2023. We thank the dissertation committee members for their insightful comments: Kaan Ozbay, Zhibin Chen, and external member Ilya Ryzhov.# REFERENCES

1. Abdulhai, B., Pringle, R., & Karakoulas, G.J. (2003). Reinforcement learning for true adaptive traffic signal control. *Journal of Transportation Engineering, 129(3)*, pp.278-285.
2. Allahviranloo, M. and Chow, J.Y., 2019. A fractionally owned autonomous vehicle fleet sizing problem with time slot demand substitution effects. *Transportation Research Part C: Emerging Technologies, 98*, pp.37-53.
3. An, K., & Lo, H.K. (2015). Robust transit network design with stochastic demand considering development density. *Transportation Research Part B: Methodological, 81*, 737-754.
4. An, K., & Lo, H.K. (2016). Two-phase stochastic program for transit network design under demand uncertainty. *Transportation Research Part B: Methodological, 84*, 157-181.
5. Auer, P., Cesa-Bianchi, N., & Fischer, P. (2002). Finite-time analysis of the multiarmed bandit problem. *Machine learning, 47(2)*, 235-256.
6. Baaj, M.H. and Mahmassani, H.S., 1995. Hybrid route generation heuristic algorithm for the design of transit networks. *Transportation Research Part C: Emerging Technologies, 3*(1), pp.31-50.
7. Baxter, M., Elgindy, T., Ernst, A. T., Kalinowski, T., & Savelsbergh, M. W. (2014). Incremental network design with shortest paths. *European Journal of Operational Research*, 238(3), 675-684.
8. Borndörfer, R., Grötschel, M., Pfetsch, M.E. (2007). A column-generation approach to line planning in public transport. *Transportation Science, 41(1)*, 123-132.
9. Bubeck, S., & Cesa-Bianchi, N. (2012). Regret analysis of stochastic and nonstochastic multi-armed bandit problems. *Foundations and Trends® in Machine Learning, 5(1)*, 1-122.
10. Bunch, D.S., Bradley, M., Golob, T.F., Kitamura, R. and Occhiuzzo, G.P., 1993. Demand for clean-fuel vehicles in California: a discrete-choice stated preference pilot project. *Transportation Research Part A: Policy and Practice, 27*(3), pp.237-253.
11. Canca, D., De-Los-Santos, A., Laporte, G., & Mesa, J.A. (2017). An adaptive neighborhood search metaheuristic for the integrated railway rapid transit network design and line planning problem. *Computers & Operations Research, 78*, 1-14.
12. Caros, N. S., & Chow, J. Y. (2021). Day-to-day market evaluation of modular autonomous vehicle fleet operations with en-route transfers. Transportmetrica B: Transport Dynamics, 9(1), 109-133.
13. Castle Labs, 2022a. The Optimal Learning Calculator, URL: http://optimallearning. princeton.edu/software/KnowledgeGradient_IndependentNormal.xlsx. Accessed on July 30, 2022.
14. Castle Labs, 2022b. Matlab implementation of the knowledge gradient for correlated beliefs using a lookup table belief model, URL: http://optimallearning.princeton.edu/ software/KGCB.zip. Accessed on July 30, 2022.
15. Cats, O., & West, J. (2020). Learning and adaptation in dynamic transit assignment models for congested networks. *Transportation Research Record, 2674(1)*, pp.113-124.
16. Ceder, A. and Israeli, Y., 1998. User and operator perspectives in transit network design. *Transportation Research Record, 1623*(1), pp.3-7.
26